\documentclass{article} 
\usepackage{iclr2020_conference,times}

\usepackage{amsmath,amsfonts,bm}

\def\eqref#1{equation~\ref{#1}}

\def\1{\bm{1}}

\DeclareMathAlphabet{\mathsfit}{\encodingdefault}{\sfdefault}{m}{sl}
\SetMathAlphabet{\mathsfit}{bold}{\encodingdefault}{\sfdefault}{bx}{n}

\usepackage{hyperref}
\usepackage{url}
\usepackage[linesnumbered, ruled, vlined]{algorithm2e}
\usepackage{graphicx}
\usepackage{enumitem}
\usepackage{subcaption}
\usepackage{makecell}

\title{Stochastic Weight Averaging in Parallel: \\ Large-Batch Training That Generalizes Well}

\author{Vipul Gupta\thanks{Equal contribution} \thanks{Work done during an internship at Apple Inc.} \\
	vipul\_gupta@berkeley.edu \\
	Deparment of EECS, UC Berkeley 
	\And Santiago Akle Serrano \footnotemark[1] \\
	sakle@apple.com \\
	Apple Inc. 
	\And Dennis DeCoste \\
	ddecoste@apple.com \\
	Apple Inc.}

\iclrfinalcopy 
\begin{document}

\maketitle

\vspace{-0.2in}

\begin{abstract}
 We propose Stochastic Weight Averaging in Parallel (SWAP), an algorithm to
 accelerate DNN training. Our algorithm uses large mini-batches to compute an
 approximate solution quickly and then refines it by averaging the weights of
 multiple models computed independently and in parallel. The resulting models
 generalize equally well as those trained with small mini-batches but are
 produced in a substantially shorter time. We demonstrate the reduction in
 training time and the good generalization performance of the resulting models on
 the computer vision datasets CIFAR10, CIFAR100, and ImageNet.
\end{abstract}

\vspace{-0.2in}

\section{Introduction}
\vspace{-0.1cm}

Stochastic gradient descent (SGD) and its variants are the de-facto methods to
train deep neural networks (DNNs). Each iteration of SGD computes an estimate
of the objective's gradient by sampling a \emph{mini-batch} of the available
training data and computing the gradient of the loss restricted to the sampled
data. 
A popular strategy to accelerate DNN training is to increase the mini-batch
size together with the available computational resources. Larger
mini-batches produce more precise gradient estimates; these allow for higher
learning rates and achieve larger reductions of the training loss per
iteration. In a distributed setting, multiple nodes can compute gradient
estimates simultaneously on disjoint subsets of the mini-batch and produce a
consensus estimate by averaging all estimates, with one
synchronization event per iteration. Training with larger mini-batches
requires fewer updates, thus fewer synchronization events, yielding good overall
scaling behavior.

Even though the training loss can be reduced more efficiently, there
is a maximum batch size after which the resulting model tends to have
worse generalization performance 
\citep{openai,keskar,hoffer17,LBempGolmant:2018,LBnotEmp:2018shallue}. This
phenomenon forces practitioners to use batch sizes below those that achieve the 
maximum throughput and limits the usefulness of large-batch training strategies.
 
Stochastic Weight Averaging (SWA) \citep{swa} is a method
that produces models with good generalization performance by averaging the
\emph{weights} of a set of models sampled from the final stages of a training
run. As long as the models all lie in a region where the population loss is
mostly convex, the average model can behave well, and in practice, it does. 

We have observed that if instead of sampling multiple models from a sequence
generated by SGD, we generate multiple independent SGD sequences and average
models from each, the resulting model achieves similar generalization performance. 
Furthermore, if all the independent
sequences use small-batches, but start from a model trained with large-batches, the resulting model
achieves generalization performance comparable
with a model trained solely with small-batches. Using these observations, we
derive \emph{Stochastic Weight Averaging in Parallel} (SWAP): A simple
strategy to accelerate DNN training by better utilizing available compute
resources. Our algorithm is simple to implement, fast and produces good results
with minor tuning. 

For several image classification tasks on popular computer vision datasets
(CIFAR10, CIFAR100, and ImageNet), we show that SWAP achieves generalization
performance comparable to models trained with small-batches but does so in time
similar to that of a training run with large-batches.  We use SWAP on some of
the most efficient publicly available models to date, and show that it's able
to substantially reduce their training times.
Furthermore, we are able to beat the state of the art for CIFAR10 and train in
$68\%$ of the time of the winning entry of the DAWNBench
competition.\footnote{The https://dawn.cs.stanford.edu/benchmark/} 

\section{Related Work}


The mechanism by which the training batch size affects the generalization
performance is still unknown. A popular explanation is
that because of the reduced noise, a model trained using larger mini-batches is
more likely to get stuck in a sharper global minima. In \citep{keskar}, the
authors argue that sharp minima are sensitive to variations in the data because
slight shifts in the location of the minimizer will result in large increases
in average loss value. However, if flatness is taken to be the curvature as
measured by the second order approximation of the loss, then counterexamples
exist.  In \citep{bengio2017sharp}, the authors transform a flat minimizer into
a sharp one without changing the behavior of the model, and in
\citep{li2018visualizing}, the authors show the reverse behavior when
weight-decay is not used.

In \citep{openai}, the authors predict that the batch size can be increased up to
a \emph{critical size} without any drop in accuracy and empirically validate
this claim. For example, the accuracy begins to drop for image classification
on CIFAR10 when the batch sizes exceed 1k samples. They postulate that when the batch size is large, the mini-batch
gradient is close to the full gradient, and further increasing the batch size
will not significantly improve the signal to noise ratio. 

In \citep{hoffer17}, the authors argue that, for a fixed number of epochs, using a 
larger batch size implies fewer model updates. They argue that changing the number 
of updates impacts the distance the weights travel away from their initialization 
and that this distance determines the generalization performance. They show
that by training with large-batches for longer times (thus increasing the number of
updates), the generalization performance of the model is recovered. Even though this
large-batch strategy generates models that generalize well, it does so in more 
time than the small-batch alternative.

Irrespective of the generalization performance, the batch size also affects the
optimization process. In \citep{LBtheoryMa:2017}, the authors show that
for convex functions in the over-parameterized setting, there is a critical batch size below which an iteration
with a batch size of $M$ is roughly equivalent to $M$ iterations with a batch
size of one, and batch-sizes larger than $M$ do not improve the rate of convergence.

Methods which use adaptive batch sizes exist
\citep{adabatch,priya2017,imagenet4mins:2018,smith2017,you2017scalingImagenet32k}.
However, most of these methods are either designed for specific datasets or
require extensive hyper-parameter tuning. Furthermore, they ineffectively use
the computational resources by reducing the batch size during part of the
training.  

Local SGD
\citep{zhang_local_sgd,stich_local_sgd,li_federated_learning,yu_parallel_restarted_sgd}
is a distributed optimization algorithm that trades off gradient precision with
communication costs by allowing workers to independently update their models for a
few steps before synchronizing. Post-local SGD \citep{post_local_sgd} is a
variant, which refines the output of large-batch training with local-SGD. The
authors have observed that the resulting model has better generalization than
the model trained with large-batches and that their scheme achieves significant
speedups. In this manner Post-local SGD is of a very similar vein than the
present work. However, while Post-local SGD lets the models diverge for $T$
iterations where $T$ is in the order of tens, SWAP averges the models once
after multiple epochs. For example, in our Imagenet exeperiments (see Sec.
\ref{sec:experiments}) we average our models after tens of thousands of
updates, while Post-local SGD does after at most 32. Because of this
difference, we believe that the mechanisms that power the success of SWAP and
Post-local SGD must be different and point to different phenomena in DNN
optimization.

Stochastic weight averaging (SWA) \citep{swa} is a method where models are
sampled from the later stages of an SGD training run. When the weights of these
models are averaged, they result in a model with much better generalization
properties.  This strategy is very effective and has been adopted in multiple
domains: deep reinforcement learning \citep{swa_rl}, semi-supervised learning
\citep{swa_semiML}, Bayesian inference \citep{swa_bayesian}, low-precision
training \citep{swa_lp}.  In this work, we adapt SWA to accelerate DNN training.

\section{Stochastic weight averaging in parallel}
\vspace{-0.1in}

We describe SWAP as an algorithm in three phases (see Algorithm
\ref{alg:SWAP}): In the first phase, all workers train a single
model by computing large mini-batch updates. Synchronization between workers is
required at each iteration and a higher learning rate is used. In the second
phase, each worker independently refines its copy of the model to produce a
different set of weights. Workers use a smaller batch size, a lower learning
rate, and different randomizations of the data. No synchronization
between workers is required in this phase. The last phase consists of averaging the weights
of the resulting models and computing new batch-normalization statistics to
produce the final output.

Phase 1 is terminated before the training loss reaches zero or the training accuracy
reaches $100\%$ (for example, a few percentage points below $100\%$). We believe that
stopping early precludes the optimization from getting stuck at a location
where the gradients are too small and allows the following stage to improve the
generalization performance. However, the optimal stopping accuracy is a
hyper-parameter that requires tuning. 

During phase 2, the batch size is appropriately reduced and small-batch
training is performed independently and simultaneously. Here, each worker (or a subset of them)
performs training using all the data, but sampling in different
random order. Thus, after the end of the training process, each worker (or subset) will have produced 
a different model.

\begin{figure}[h]
	\centering
\includegraphics[scale=0.18]{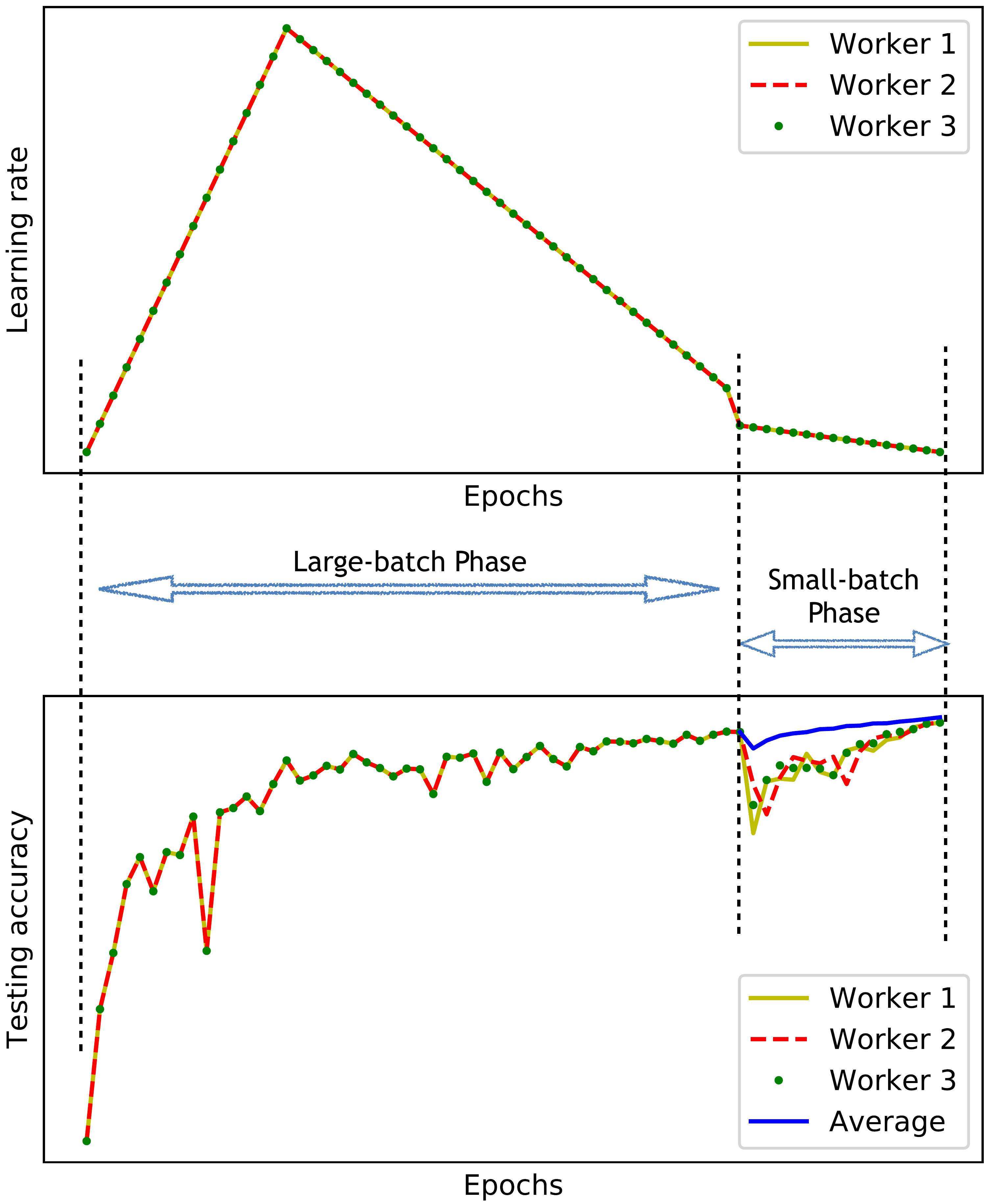} 
\vspace{-0.1in}
\caption{\small Learning rate schedules and CIFAR10 test accuracies for workers participating in SWAP. 
	The large-batch phase with synchronized models is followed by the small-batch phase with
	diverging independent models. The test accuracy of the averaged weight model is
	computed by averaging the independent models and computing the test loss for the resulting model.}
	\label{fig:swa_lr}
	\vspace{-0.2in}
\end{figure}

Figure \ref{fig:swa_lr} plots the accuracies and learning-rate schedules for a 
run of SWAP. During the large-batch phase (phase 1), all workers share a common
model and have the same generalization performance. During the small-batch phase
(phase 2) the learning rates for all the workers are the same but their testing
accuracies differ as the stochasticity causes the models to diverge from each
other. We also plot the test-accuracy of the averaged model that would result
were we to stop phase 2 at that point. Note that the averaged model performs
consistently better than \emph{each individual model}.

\begin{algorithm}[t]
\SetAlgoLined
Number of workers $W$; Weight initialization $\theta_0$; $t = 0$\\ 
Training accuracy, $\tau$, at which to exit phase one\\
Learning rate schedules $LR_1$ and $LR_2$ for phase one and two, respectively\\
Mini-batch sizes $B_1$ and $B_2$ for phase one and two, respectively \\
Gradient of loss function for sample $i$ at weight $\theta$: $g^i$\\
\newcommand{\Update}{\operatorname{SGDUpdate}}
$\Update(\cdot):$ A function that updates the weights using SGD with momentum and weight decay \\
\textbf{Phase 1:}\\
\While{Training accuracy $\leq \tau$}{
	 $\eta \gets LR_1(t)$ \\
	 \For{$w$ in $[0,...,W-1]$ In parallel}{	 
	    $B^w \gets$ random sub-sample of training data with size $\frac{B_1}{W}$ \\\
	    $g^w \gets \frac{W}{|B_1|}\sum_{i \in B^w} g^i$ worker gradient
  	 }
	 $ g_t \gets \frac{1}{W}\sum g^w$ synchronization of worker gradients \\	
	 $\theta_{t+1} \gets \theta_t + \Update(\eta_t, g_t, g_{t-1},\cdots) $ \tcc*{first order method update} 
    $~t = t+1$;  $T = t$ \\
}

\textbf{Phase 2:}\\

\For{$t$ in $[T,T+Q]$}{
	$\eta \gets LR_2(t-T)$ \\
 	\For{$w$ in $[0,...,W-1]$ In parallel}{ 
		$B^w \gets$ random sub-sample of training data with size $B_2$ \\
	    	$g^w \gets \frac{1}{|B_2|}\sum_{i \in B^w} g^i$ worker gradient \\
	 	$\theta^w_{t+1} \gets \theta^w_t + \Update(\eta_t, g^w_t, g^w_{t-1},\cdots) $ 
		\tcc*{first order method update at local worker}
		}
		}
\tcc{We get $W$ different models at the end of phase 2}
\textbf{Phase 3:} $\hat \theta_\ell \gets \frac{1}{W}\sum \theta_{T+Q}^i$ produce averaged model \\
  Compute batch-norm statistics for $\hat \theta_\ell$ to produce $\theta_\ell$ \\
\KwResult{Final model $\theta_\ell$}
\caption{Stochastic Weight Averaging in Parallel (SWAP)}
\label{alg:SWAP}
\end{algorithm}

\section{Loss Landscape Visualization around SWAP iterates}

To visualize the mechanism behind SWAP, we plot the error achieved by our test
network on a plane that contains the outputs of the three different phases of the
algorithm. Inspired by \citep{garipov} and \citep{swa}, we pick orthogonal vectors $u, v$
that span the plane which contains $\theta_1, \theta_2, \theta_3$. We plot the loss 
value generated by model $\theta = \theta_1 + \alpha u + \beta v$ at the location
($\alpha$, $\beta$).
To plot a loss value, we first generate a weight vector $\theta$, compute the batch-norm
statistics for that model (through one pass over the training data), and
then evaluate the test and train accuracies. 

\begin{figure}[t] 
	\centering 
	\begin{minipage}{0.43\textwidth} 
		\centering
		\includegraphics[height=1.5in]{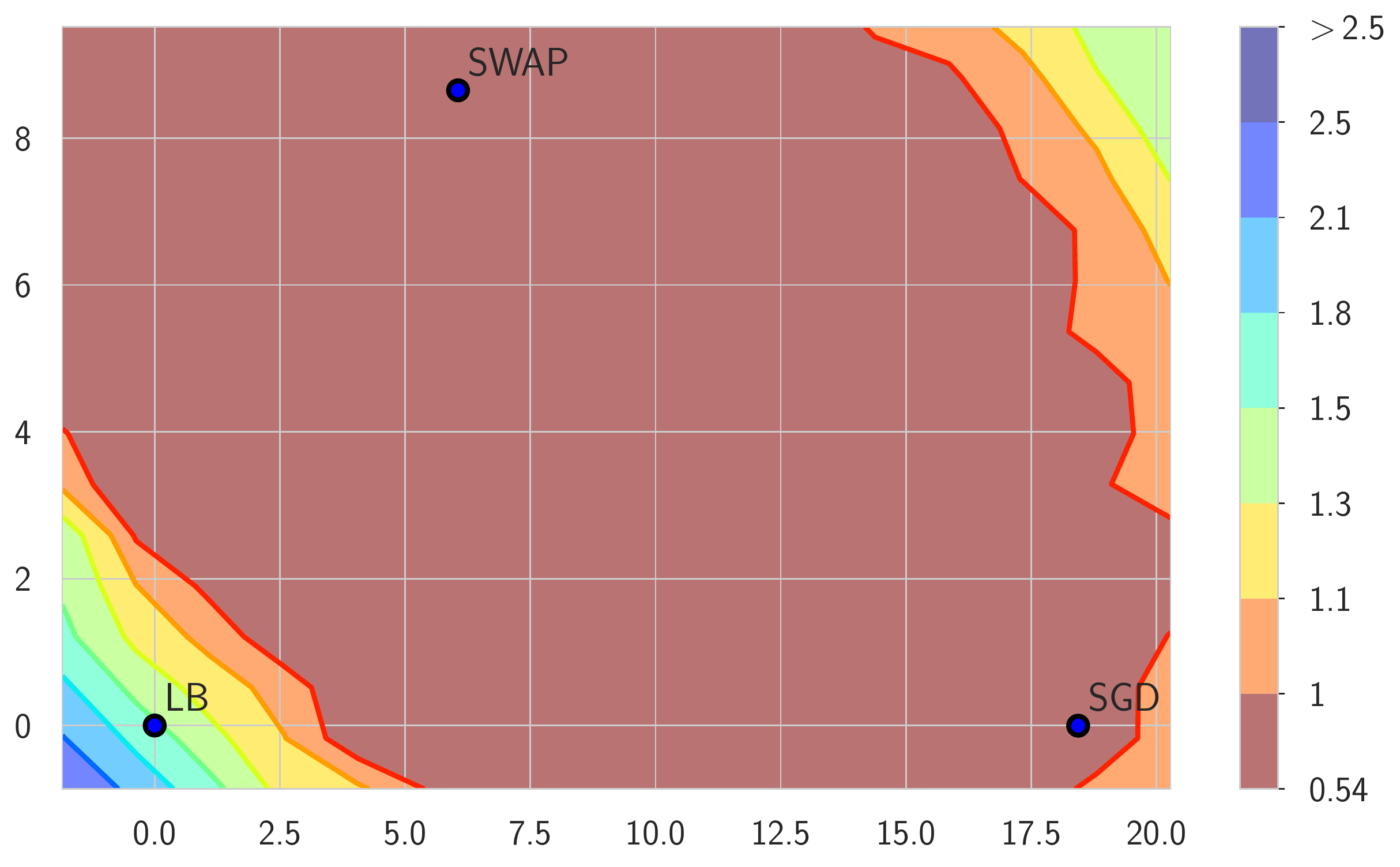} 
		\subcaption{Train Error (\%)} \label{fig:vis_lb_TrErr} 
	\end{minipage}
	\begin{minipage}{0.43\textwidth} 
		\centering
		\includegraphics[height=1.5in]{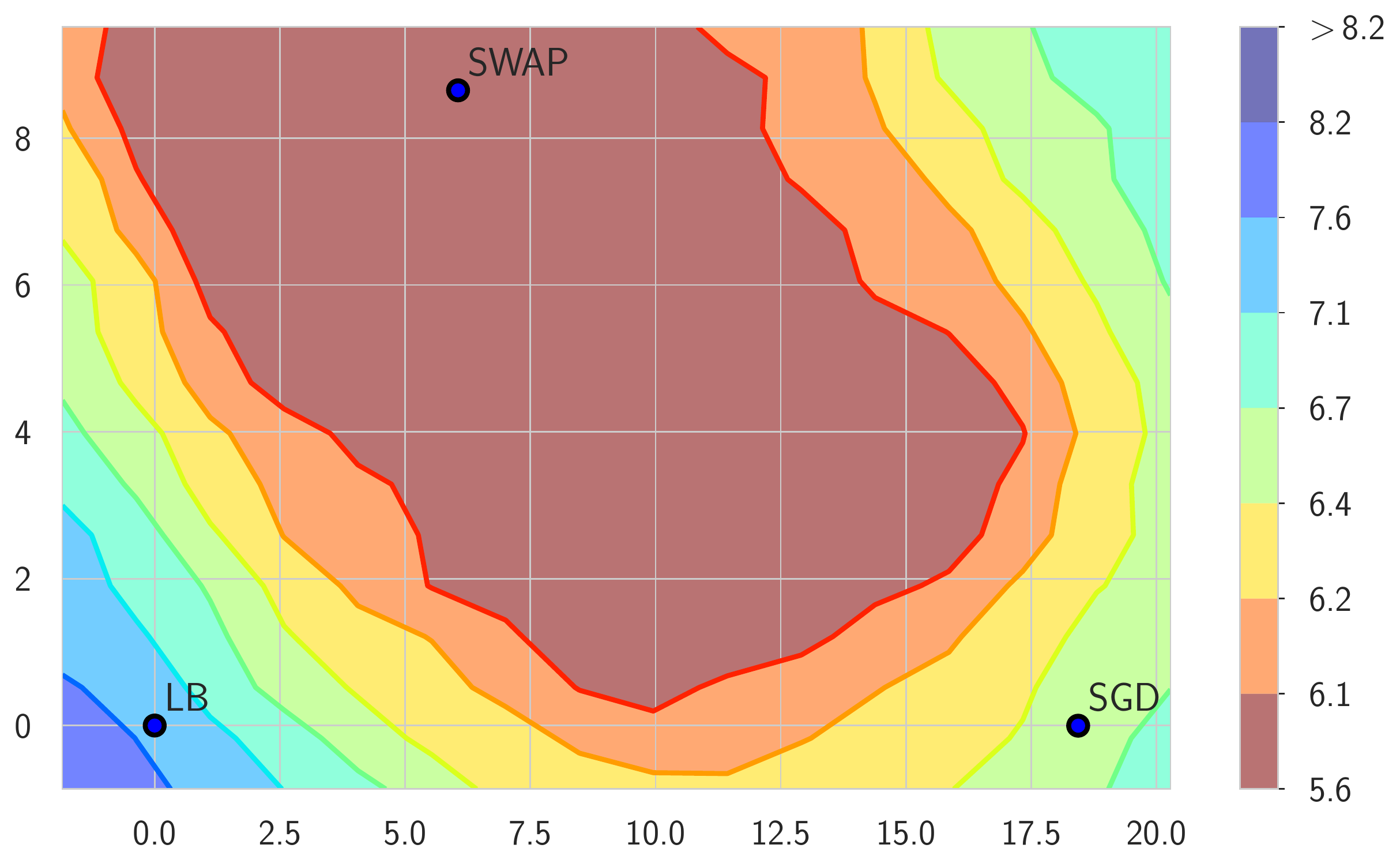} 
		\subcaption{Test Error (\%)}
		\label{fig:vis_lb_TeErr} 
	\end{minipage} 
	\vspace{-0.1in}
\caption{\small
	CIFAR10 train and test error restricted to a 2D plane spanned by the output of phase 1 (`LB'), one of the outputs of phase 2 (`SGD') and the averaged model (`SWAP').} 
	\label{fig:vis_lb} 
	\vspace{-0.2in}
\end{figure}

In Figure \ref{fig:vis_lb}, we plot the training and testing error for the
CIFAR10 dataset. Here `LB' marks the output of phase one, `SGD' the output of
a single worker after phase two, and `SWAP' the final model. Color codes
correspond to error measures at the points interpolated on the plane. 
In Figure \ref{fig:vis_lb_TrErr}, we observe that the level-sets of the training 
error (restricted to this plane) form an almost convex basin and that both the
output of phase 1 (`LB')\footnote{Recall that the weights `LB' are obtained by stopping the large-batch training early in phase 1. Hence, the training error for `LB' is worse than `SGD' and `SWAP'.} and the output of one of the workers of phase 2 (`SGD') 
lie in the outer edges of the basin. Importantly, during phase 2 the model
traversed to a \emph{different side} of the basin (and not to the center). Also, the final 
model (`SWAP') is closer to the center of the basin. 

\begin{figure}[t]
    \centering
    \begin{minipage}{0.43\textwidth}
        \centering
        \includegraphics[height=1.5in]{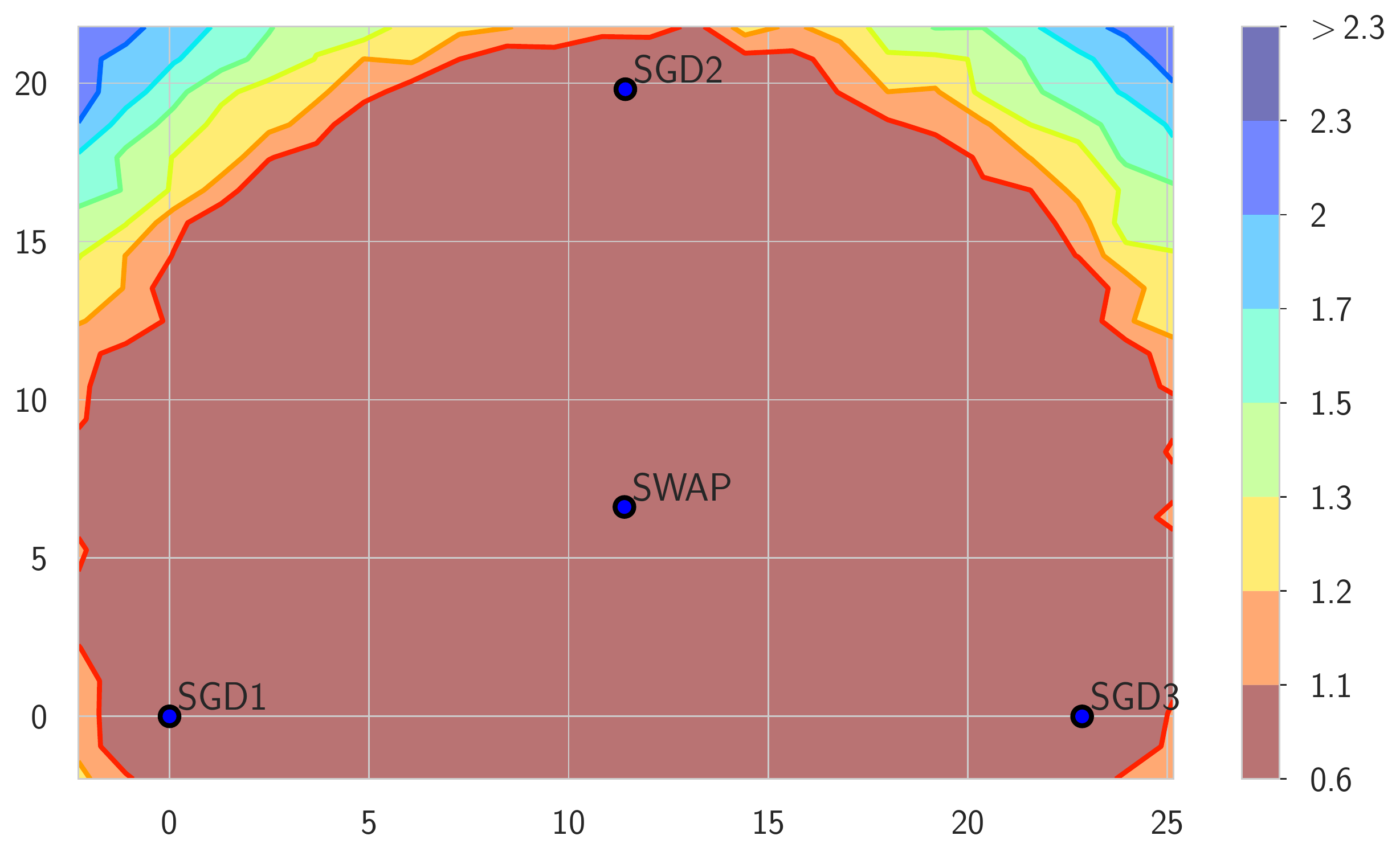}
        \subcaption{Train Error (\%) }
        \label{fig:vis_sgd_TrErr}
    \end{minipage}%
    ~ ~
    \begin{minipage}{0.43\textwidth}
        \centering
        \includegraphics[height=1.5in]{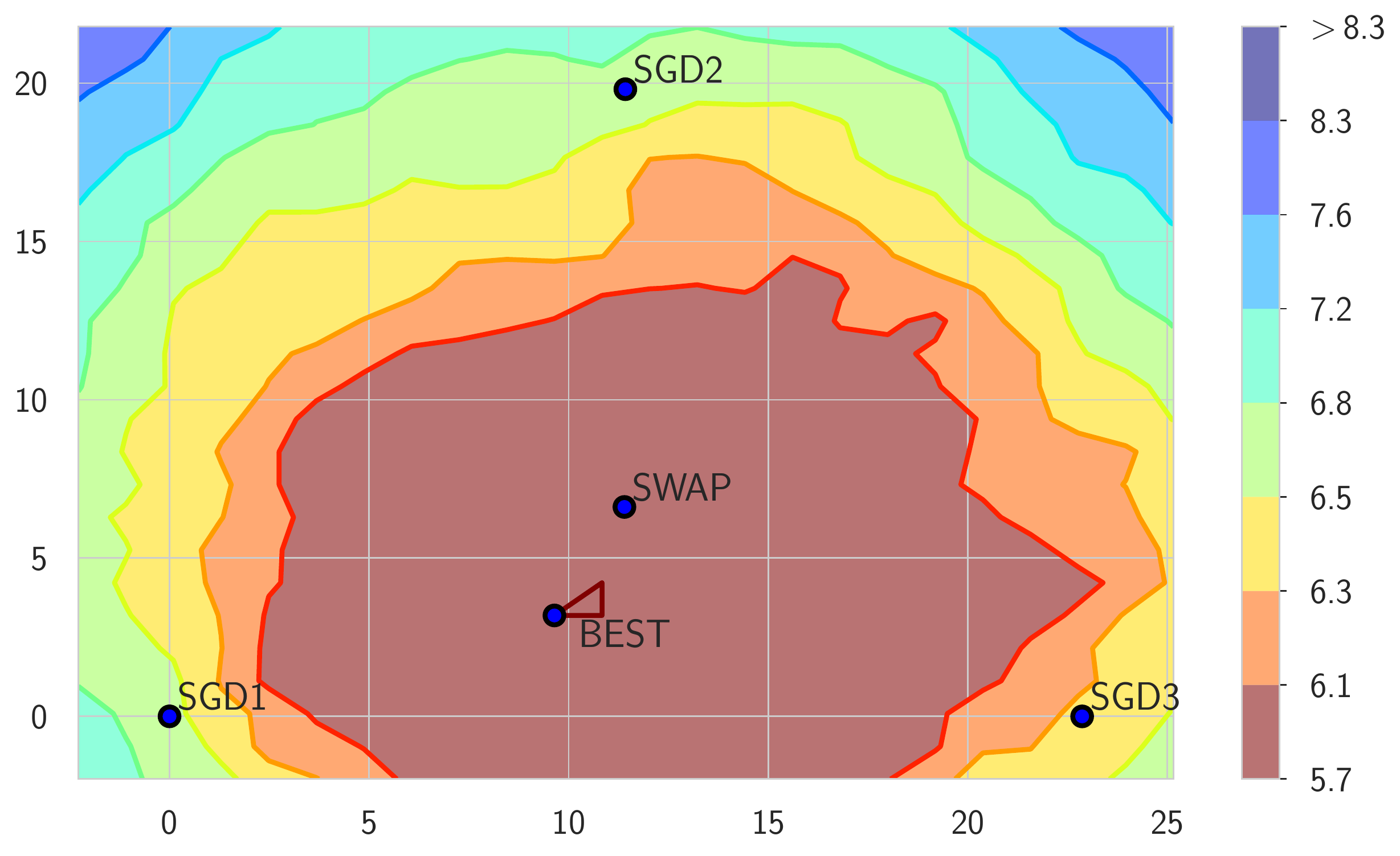}
        \subcaption{Test Error (\%) }
        \label{fig:vis_sgd_TeErr}
    \end{minipage}
\vspace{-0.1in}
    \caption{\small
    CIFAR10 train and test error restricted to a 2D plane spanned by the
    output of three workers after phase 2 (`SGD1', `SGD2', `SGD3') and location
    of the average model (`SWAP'). The minimum test error achievable for models restricted to this 
    region of the plane (marked as \textit{BEST}).}
    \label{fig:vis_sgd}
\end{figure}

When we visualize these 
three points on the test loss landscape (Figure \ref{fig:vis_lb_TeErr}), we observe
that the variations in the topology of the basin cause the `LB' and `SGD' 
points to fall in regions of higher error. But, since the `SWAP' point is closer 
to the center of the basin, it is less affected by the change in topology. 
In Figure \ref{fig:vis_sgd}, we neglect the `LB' point and plot the plane spanned
by three workers `SGD1', `SGD2', `SGD3'. In Figure \ref{fig:vis_sgd_TrErr}, we
can observe that these points lie at different sides of the training error
basin while `SWAP' is closer to the center. In Figure \ref{fig:vis_sgd_TeErr},
we observe that the change in topology causes the worker points to lie in
regions of higher testing errors than `SWAP', which is again close to the
center of both basins. For reference, we have also plotted the best model that
can be generated by this region of the plane.

\vspace{-0.1in}
\subsection{Sampling from independent runs of SGD or sampling from one}
\vspace{-0.05in}

In \citep{mandt2017sgd}, the authors argue that in the later stages of SGD
the weight iterates behave similar to an Ornstein Uhlenbeck process. So, by
maintaining a constant learning rate the SGD iterates should reach a stationary
distribution that is similar to a high-dimensional Gaussian. This distribution
is centered at the local minimum, has a covariance that grows proportionally
with the learning rate, inversely proportional to the batch size and has a
shape that depends on both the Hessian of the mean loss and covariance of the
gradient.

The authors of \citep{swa} argue that by virtue of being a high dimensional Gaussian all
the mass of the distribution is concentrated near the `shell' of the ellipsoid, and
therefore, it is unlikely for SGD to access the interior. They further argue
that by sampling weights from an SGD run (leaving enough time steps between
them) will choose weights that are spread out on the surface of this ellipsoid
and their average will be closer to the center. 

Without any further assumptions, we can justify sampling from different SGD
runs (as done in phase 2 during SWAP). As long as all runs start in the same basin of attraction, and provided
the model from \citep{mandt2017sgd} holds, all runs will converge to the same
stationary distribution, and each run can generate independent samples
from it.

\vspace{-0.1in}
\subsection{Orthogonality of the gradient and the direction to the center of basin}
\vspace{-0.05in}

\begin{figure}[t] 
\vspace{-0.1in}
\centering 
	\includegraphics[height=1.4in]{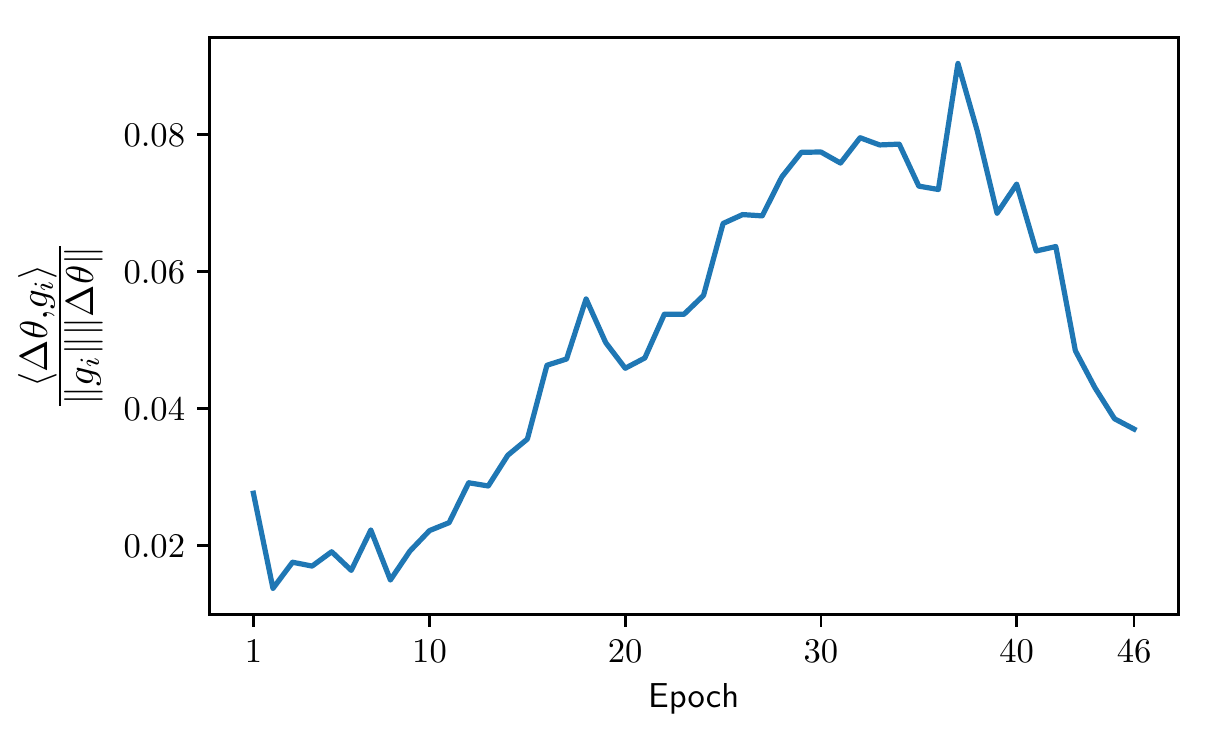} 
	\vspace{-0.1in}
	\caption{\small Cosine similarity between direction of gradient descent and $\Delta \theta$ } 
	\label{fig:grad_comp}
	\vspace{-0.2in}
\end{figure}

To win some intuition on the advantage that SWA and SWAP have over SGD, we measure
the cosine similarity between the gradient descent direction, $-g_i$, and the
direction towards the output of SWAP, $\Delta\theta =
\theta_{\text{swap}}-\theta_i$.  In Figure \ref{fig:grad_comp}, we see that the
cosine similarity, $\frac{\langle \Delta\theta, -g_i \rangle}{\|g_i\| \|\Delta
\theta\|}$, decreases as the training enters its later stages. We believe that
towards the end of training, the angle between the gradient direction and the
directions toward the center of the basin is large, therefore the process moves
mostly orthogonally to the basin, and progress slows. However,
averaging samples from different sides of the basin can (and does) make faster
progress towards the center. 


\vspace{-0.1in}
\section{Experiments} \label{sec:experiments}
\vspace{-0.1in}
In this section we evaluate the performance of SWAP for image classification tasks on the
CIFAR10, CIFAR100, and ImageNet datasets. 

\subsection{CIFAR10 and CIFAR100}
\label{sec:cifar_exp} 
\vspace{-0.1in}
For the experiments in this subsection, we found the best
hyper-parameters using grid searches (see Appendix \ref{sec:appendix} for details). We train using mini-batch SGD with Nesterov
momentum (set to 0.9) and weight decay of $5\times10^{-4}$. We augment the data using cutout
\citep{cutout} and use a fast-to-train custom ResNet 9 from a
submission \footnote{https://github.com/davidcpage/cifar10-fast} to the
DAWNBench leaderboard \citep{dawn}. All experiments were run on one machine
with 8 NVIDIA Tesla V100 GPUs and use Horovod \citep{hvd} to distribute the
computation. All statistics were collected over 10 different runs.


\textbf{CIFAR10}: For these experiments, we used the following settings---SWAP
phase one: 4096 samples per batch using 8 GPUs (512 samples per GPU).  Phase
one is terminated when the training accuracy reaches $98\%$ (on average 108
epochs).  SWAP phase two: 8 workers with one GPU each and 512 samples per batch
for 30 epochs.  The experiment that uses only large-batches had
4096 samples per batch across 8 GPUs and is run for 150 epochs. The experiments
that use only small-batches had 512 samples per batch on 2 GPUs and is trained
for 100 epochs.

Table \ref{table:cifar10} compares the best test accuracies and corresponding
training times for models trained with small-batch only, with large-batch only,
and with SWAP. We report the average accuracy of the workers before averaging
and the accuracy of the final model. 

\begin{table}[h] \centering \begin{tabular}{ccc} \multicolumn{1}{c}{\bf
		CIFAR10}  &\multicolumn{1}{c}{\bf Test Accuracy (\%)} &
		\multicolumn{1}{c}{\bf Training Time (sec)}\\
 \hline SGD (small-batch) & $95.24 \pm 0.09$ & $254.12 \pm 0.62$ \\ SGD
 (large-batch) & $94.77 \pm 0.23$ & $132.62 \pm 1.09$ \\ SWAP (before
 averaging) & $94.70 \pm 0.20$ & $167.57 \pm 3.25$ \\ SWAP (after averaging) &
 $95.23 \pm 0.08$ & $169.20 \pm 3.25$ \\ \hline \end{tabular} \caption{Training
 Statistics for CIFAR10} \label{table:cifar10} 
\vspace{-0.1in}
 \end{table}

\textbf{CIFAR100}: For these experiments, we use the following settings---SWAP
phase one: 2048 samples per batch using 8 GPUs (256 samples per GPU). Phase one
exits when the training accuracy reaches $90\%$ (on average 112 epochs).  SWAP
phase two: 8 workers with one GPU each and 128 samples per batch, training for for
10 epochs. The experiments that use only large-batch training were run for 150
epochs with batches of 2048 on 8 GPUs 
The experiments that use only small-batch were trained for 150 epochs using batches of 128 on 1 GPU.

\begin{table}[h] \centering \begin{tabular}{ccc} \multicolumn{1}{c}{\bf
		CIFAR100}  &\multicolumn{1}{c}{\bf Test Accuracy (\%)} &
		\multicolumn{1}{c}{\bf Training Time (sec)}\\ \hline SGD
		(small-batch) & $77.01 \pm 0.25$ & $573.76 \pm 2.25$ \\ SGD
		(large-batch) & $75.84 \pm 0.35$ & $116.13 \pm 1.35$ \\ SWAP
		(before averaging) & $75.74 \pm 0.15$ & $123.11 \pm 1.85$ \\
		SWAP (after averaging) & $78.18 \pm 0.21$ & $125.34 \pm 1.85$
		\\ \hline \end{tabular} \caption{Training Statistics for
	CIFAR100} \label{table:cifar100} 
\vspace{-0.1in}
	\end{table}

Table \ref{table:cifar100} compares the best test accuracies and corresponding
training times for models trained with only small-batches (for 150 epochs),
with only large-batches (for 150 epochs), and with SWAP. For SWAP, we report
test accuracies obtained using the last SGD iterate before averaging, and
test accuracy of the final model obtained after averaging. We observe significant
improvement in test accuracies after averaging the models.

 For both CIFAR 10 and CIFAR100, training with small-batches achieves higher
 testing accuracy than training with large-batches but takes much longer
 to train. SWAP, however, terminates in time comparable
 to the large-batch run but achieves accuracies on par (or better) than small
 batch training.

\textbf{Achieving state of the art training speeds for CIFAR10}: 
At the time of writing the front-runner of the DAWNBench competition 
takes 37 seconds with 4 Tesla V100 GPUs to train CIFAR10 to $94\%$ test accuracy.
Using SWAP with 8 Tesla V100 GPUs, 
a phase one batch size of 2048 samples and 28 epochs, and a phase two batch size 
of 256 samples for one epoch is able to reach the same accuracy in 27 seconds.

\subsection{Experiments on ImageNet}

\begin{figure}[h] 
\centering 
\begin{minipage}{0.42\textwidth} 
	\centering
	\includegraphics[height=1.5in]{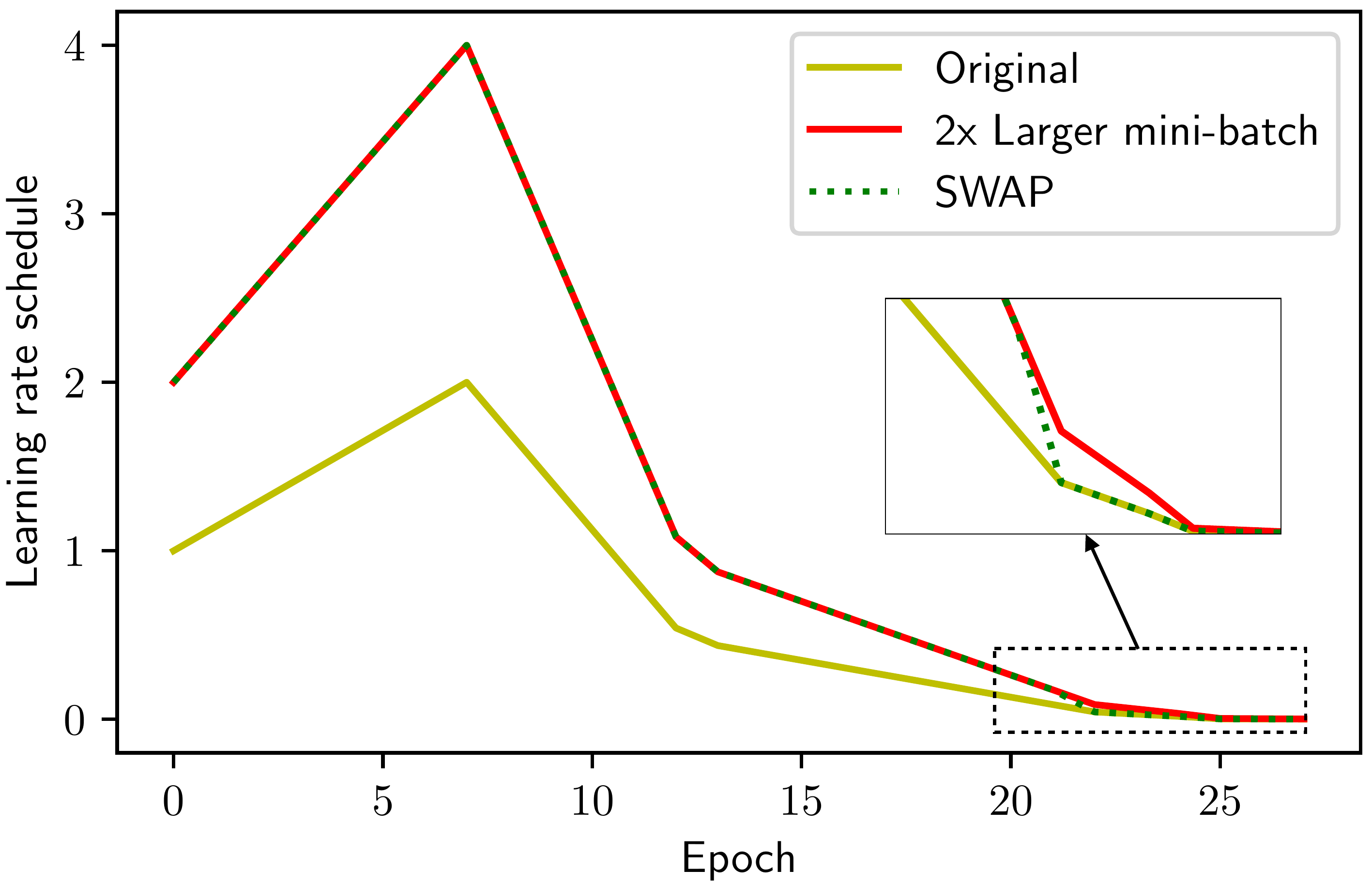} \subcaption{Learning rate schedule} 
\label{fig:imgnet_lr} 
\end{minipage} ~ 
\begin{minipage}{0.42\textwidth}
	\centering 
\includegraphics[height=1.55in]{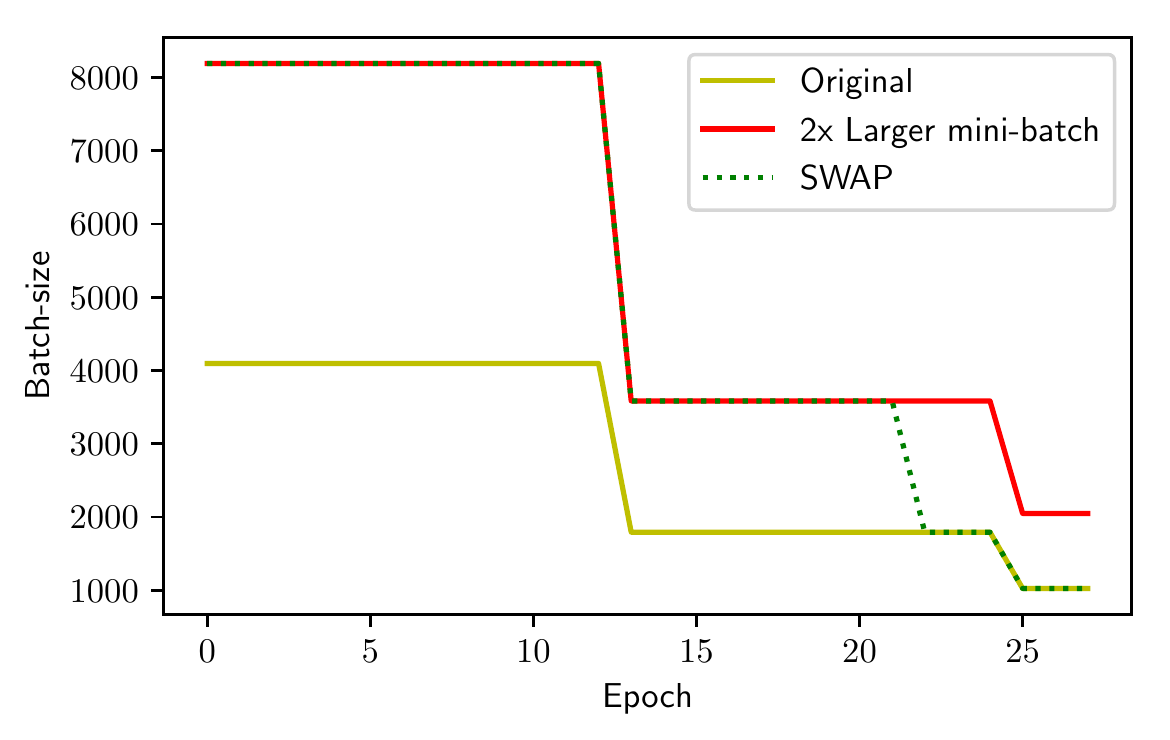} 
\subcaption{Batch sizes across epochs for ImageNet} 
\label{fig:imgnet_bs}
\end{minipage} 
\vspace{-0.1in}
\caption{\small
Learning rate and mini-batch schedules used for ImageNet.  
	The original schedule for 8 GPUs was taken from an existing DAWNBench submission.
	For a larger batch experiment, we double the batch size, double the number of GPUs and double the learning rate of the original schedule.
	For SWAP, we switch from the modified schedule to the original schedule as we move from phase 1 to phase 2.}
\label{fig:imgnet} 
\vspace{-0.1in}
\end{figure}

We use SWAP to accelerate a publicly available fast-to-train ImageNet model
with published learning rate and batch size schedules \footnote{Available at
https://github.com/cybertronai/imagenet18\_old}. The default settings for this
code modify the learning-rates and batch sizes throughout the optimization (see
Figure \ref{fig:imgnet}). Our small-batch experiments train ImageNet for 28 epochs using
the published schedules with no modification and are run on 8 Tesla V100 GPUs.
Our large-batch experiments modify the schedules by doubling the batch size and
doubling the learning rates (see Figure \ref{fig:imgnet}) and are run on 16
Tesla V100 GPUs. For SWAP phase 1, we use the large-batch settings for 22
epochs, and for SWAP phase 2, we run two independent workers each with 8 GPUs using
the settings for small-batches for 6 epochs. 

We observe that doubling the batch size reduces the Top1 and Top5 test
accuracies with respect to the small-batch run. SWAP, however, recovers the
generalization performance at substantially reduced training times. Our results
are compiled in Table \ref{table:imagenet} (the statistics were collected over
3 runs). We believe it's worthy of mention that these accelerations were
achieved with no tuning other than increasing the learning rates proportionally
to the increase in batch size and reverting to the original schedule when
transitioning between phases. 

\begin{table}[h] \centering \begin{tabular}{cccc} 
\multicolumn{1}{c}{\bf ImageNet}  & 
\multicolumn{1}{c}{\bf Top1 Accuracy (\%)} &
\multicolumn{1}{c}{\bf Top5 Accuracy (\%)} &
\multicolumn{1}{c}{\bf Training Time (min)}\\ 
\hline 
SGD (small-batch) & $76.14\pm 0.07$ & $93.30 \pm 0.07$ & $235.29 \pm 0.33$\\ 
SGD (large-batch) & $75.86 \pm 0.03$ & $92.98 \pm 0.06$ & $127.20 \pm 0.78$ \\ 
SWAP (before averaging) & $75.96 \pm 0.02$ & $93.15 \pm 0.02$ & $149.12 \pm 0.55$\\
SWAP (after averaging) & $76.19 \pm 0.03$ & $93.32 \pm 0.02$ & $156.55 \pm 0.56$
\\ \hline 
\end{tabular} 
\caption{Training Statistics for ImageNet} 
\label{table:imagenet} 
\vspace{-0.1in}
\end{table}

\subsection{Empirical comparison of SWA and SWAP}
\vspace{-0.05in}

\begin{figure}[t] 
\centering 
\begin{minipage}{0.32\textwidth} \centering
	\includegraphics[height=1.2in]{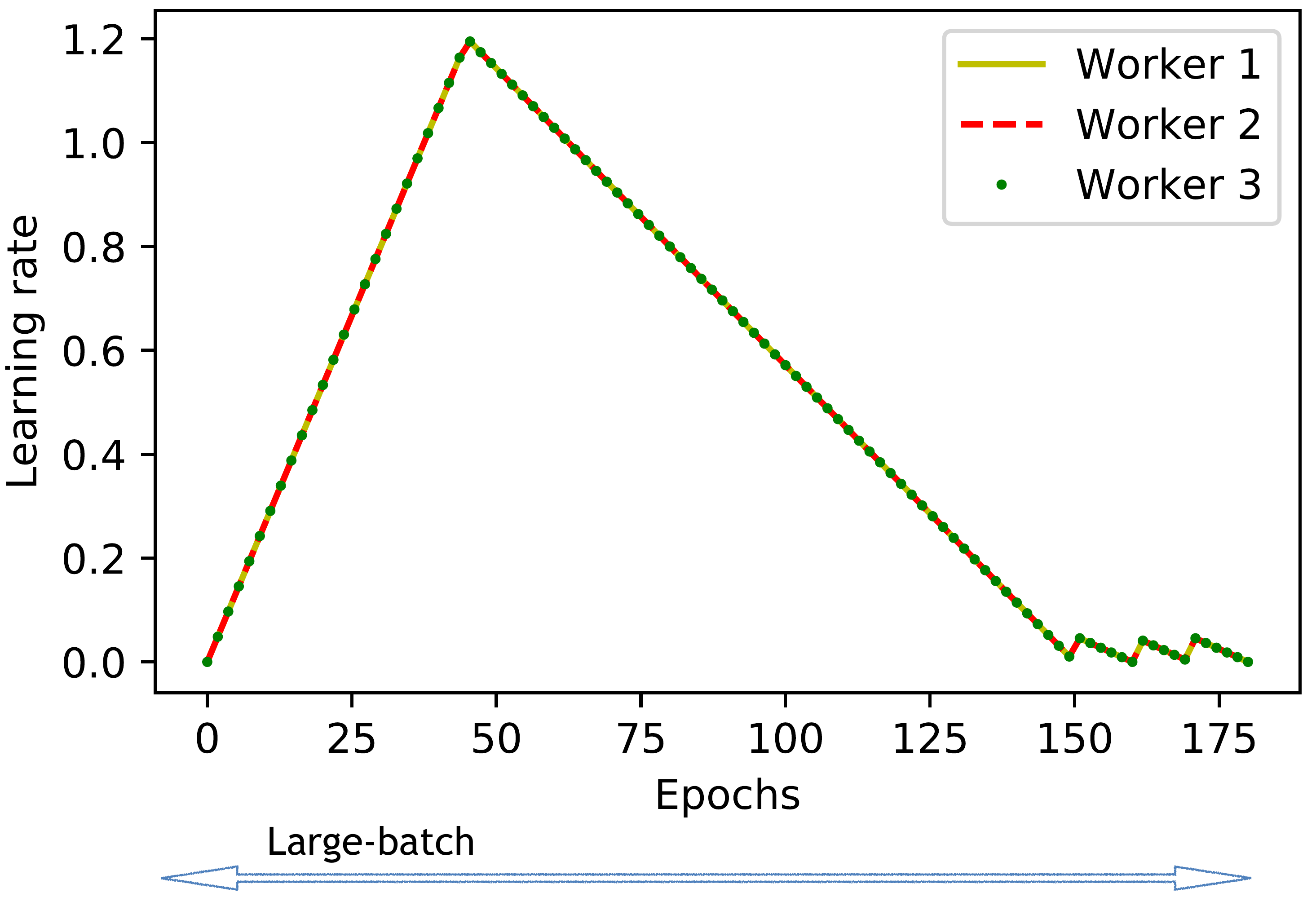} \subcaption{Large-batch SWA} 
\label{fig:swa_lb} \end{minipage} ~ 
\begin{minipage}{0.32\textwidth}
	\centering \vspace{3mm}
\includegraphics[height=1.17in]{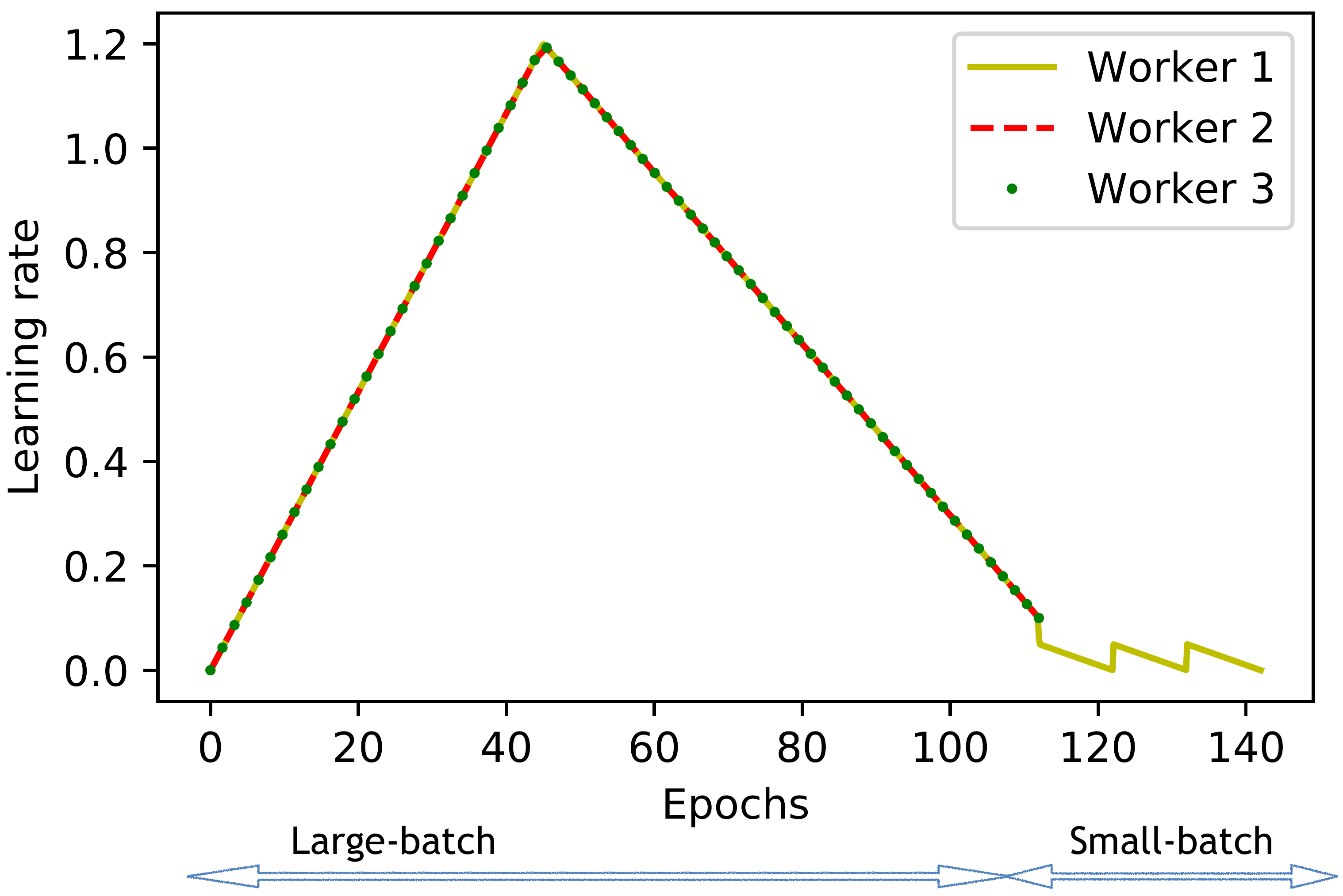} 
\subcaption{Large-batch training followed by SWA with small-batches} 
\label{fig:swa_lb_sb}
\end{minipage} 
~ \begin{minipage}{0.32\textwidth} \centering
	\includegraphics[height=1.2in]{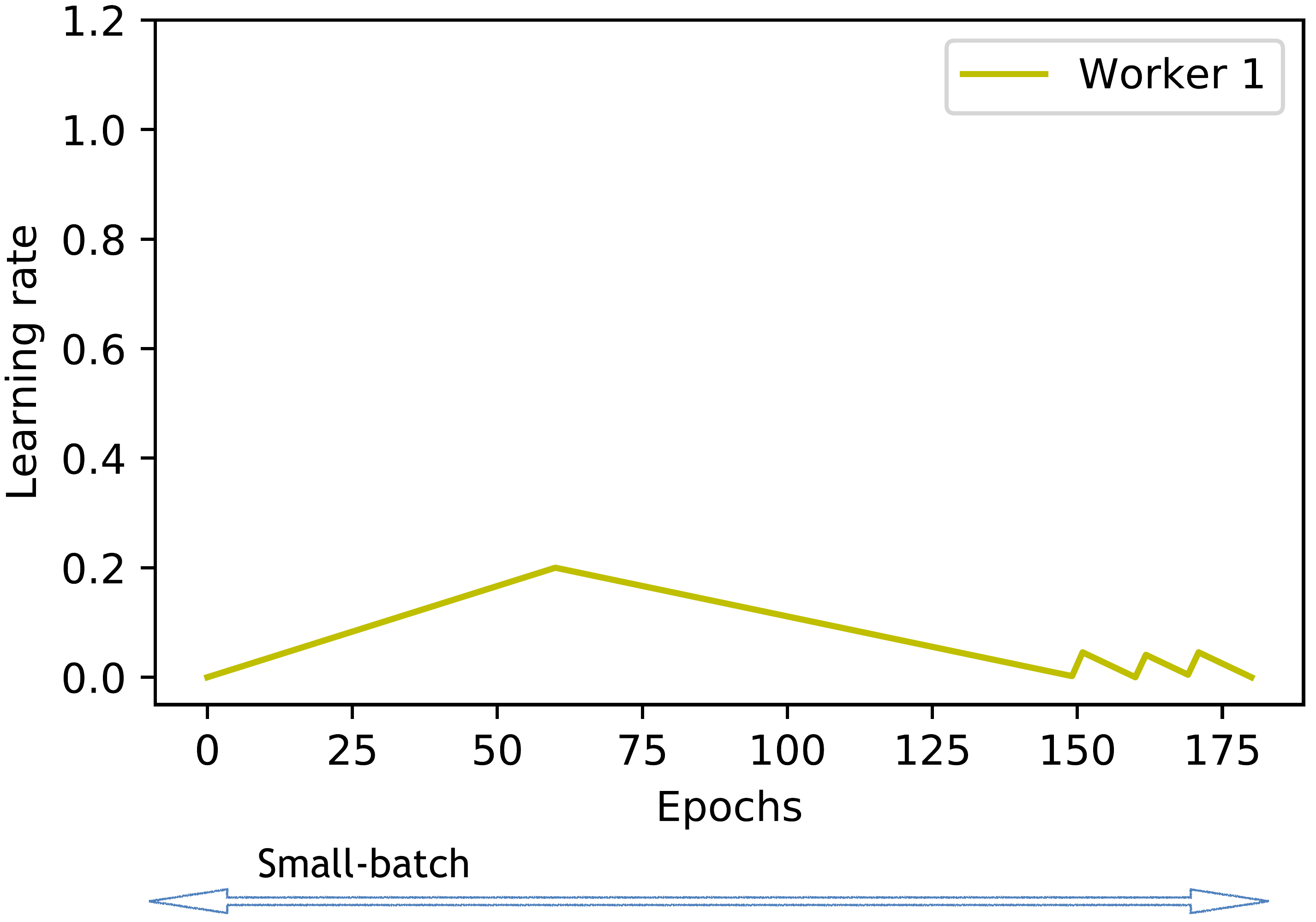} \subcaption{Small-batch
	SWA} \label{fig:swa_sb} \end{minipage} 
	\caption{\small Illustration of SWA
with different batch sizes} \label{fig:swa} 
\vspace{-0.2in}
\end{figure}{}

We now compare SWAP with SWA: the sequential weight averaging algorithm from
\cite{swa}. For the experiments in this section, we use the CIFAR100
dataset. We sample the same number of models for both SWA and SWAP and maintain
the same number of epochs per sample. For SWA, we sample each model with 10
epochs in-between and average them to get the final model. For SWAP, we run 8
independent workers for 10 epochs each and use their average as the final
model. 

\textbf{Large-batch SWA}: 
We explore if SWA can recover the test accuracy of small-batch training on a
large-batch training run. We use the same (large) batch size throughout. We
follow an initial training cycle with cyclic learning rates (with cycles of 10
epochs) to sample 8 models (one from the end of each cycle).  See
Figure \ref{fig:swa_lb} for an illustration of the learning rate schedule. 

As expected we observe that the large-batch training run achieves lower
training accuracy, but surprisingly SWA was unable to improve it 
(see Table \ref{table:swa_vs_swap}, row 1).

\textbf{Large-batch followed by small-batch SWA}: 
We evaluate the effect of executing SWA using small-batches after a large-batch
training run. We interrupt the large-batch phase at the same accuracy we
interrupt phase 1 of our CIFAR100 experiment (Table \ref{table:cifar100}). In this
case, the small-batch phase uses a single worker and samples the models
sequentially. SWA is able to reach the test accuracy of a small-batch run but
requires more than three times longer than SWAP to compute the model (see Table
\ref{table:swa_vs_swap}, row 2). An illustration of the learning rate schedule
is provided in Figure \ref{fig:swa_lb_sb}. 

\textbf{Small-batch SWA and SWAP}: 
We start the SWA cyclic learning rate
schedule from the best model found by solely small-batch training (table
\ref{table:cifar100}, row 1). Since the cycle length and cycle count are fixed,
the only free parameter is the peak learning rate. We select this using a
grid-search. Once the SWA schedule is specified, we re-use the peak learning
rate settings in SWAP. We start phase two from the model that was generated as the
output of phase 1 for the experiment on section~\ref{sec:cifar_exp} reported on
table~\ref{table:cifar100} rows 3 and 4. With these settings, small-batch SWA
achieves better accuracy than SWAP (by around $\sim0.9\%$) at 6.8x more
training time. 

Next, we wish to explore the speed-up that SWAP achieves over SWA if the 
precision of SWA is set as a target.  To that end, we relax the constraints on
SWAP. By increasing the phase two schedule from one $10$ epoch cycle to two
$20$ epoch cycles and sampling two models from each worker (16 models) the
resulting model achieved a test accuracy of $79.11\%$ in $241$ seconds or
$3.5$x less time.



\begin{table}[h] \centering \begin{tabular}{cccc} {\bf CIFAR100}
&\makecell{\bf Test accuracy \\ 
\bf before averaging (\%)} & \makecell{\bf Test accuracy after \\
\bf  averaging (\%)} & {\bf \makecell{Training\\ Time (sec)}}\\ 
\hline Large-batch SWA & $76.06 \pm 0.25$ &  $76.00\pm 0.31$ & $376.4 \pm 2.25$ \\
\makecell{Large-batch followed \\ by small-batch SWA} & $76.26
\pm 0.35$ &  $78.12 \pm 0.14$ & $398.0 \pm 1.35$ \\ 
Small-batch SWA & $76.80 \pm 0.15$  &  $79.09\pm 0.19$ & $848.6 \pm 5.61$ \\ 
SWAP (10 small-batch epochs) & $75.74 \pm 0.15$ & $78.18 \pm 0.21$ & $125.30 \pm 1.85$ \\
SWAP (40 small-batch epochs) & $76.19 \pm 0.19$ & $79.11 \pm 0.12$ & $241.54 \pm 1.62$\\
\hline 
	\end{tabular} \caption{Comparison: SWA versus SWAP}
\label{table:swa_vs_swap} 
\vspace{-0.1in}
\end{table}

\section{Conclusions and Future Work}
\vspace{-0.1in}

We propose Stochastic Weight Averaging in Parallel (SWAP), an algorithm that
uses a variant of Stochastic Weight Averaging (SWA) to improve the
generalization performance of a model trained with large mini-batches. Our
algorithm uses large mini-batches to compute an approximate solution quickly
and then refines it by averaging the weights of multiple models trained using small-batches. The final model obtained after averaging has good generalization performance and is trained in a shorter time. We believe that this variant 
and this application of SWA are novel.

We observed that using large-batches in the initial stages of training does not
preclude the models from achieving good generalization performance. That is, by
refining the output of a large-batch run, with models sampled sequentially as
in SWA or in parallel as in SWAP, the resulting model is able to perform as well
as the models trained using small-batches only. We confirm this in the image 
classification datasets CIFAR10, CIFAR100, and ImageNet. 

Through visualizations, we complement the existing evidence that averaged
weights are closer to the center of a training loss basin than the models
produced by stochastic gradient descent. It's interesting to note that the
basin into which the large mini-batch run is converging to seems to be the same
basin where the refined models are found. So, it is possible that regions with
bad and good generalization performance are connected through regions of low
training loss and, more so, that both belong to an almost convex basin.
Our method requires the choice of (at least) one more hyperparameter: the
transition point between the large-batch and small-batch. For our experiments,
we chose this by using a grid search. A principled method to choose the
transition point will be the focus of future work.

In future work we intend to explore the behavior of SWAP when used with other optimization schemes, such as Layer-wise Adaptive Rate Scaling (LARS) \citep{you2017scalingImagenet32k}, mixed-precision training \cite{imagenet4mins:2018}, post-local SGD \citep{post_local_sgd} or NovoGrad \citep{novograd}. The design of SWAP allows us to substitute any of these for the large-batch stage, for example, we can use local SGD to accelerate the first stage of SWAP by reducing the communication overhead.

\newpage


\bibliography{iclr2020_conference} \bibliographystyle{iclr2020_conference}

\newpage

\appendix
\section{Hyperparameters for CIFAR10 and CIFAR100 Experiments}
\label{sec:appendix}
We provide the parameters used in the experiments of Section
\ref{sec:cifar_exp}. These were obtained by doing independent grid searches for
each experiment. For all CIFAR experiments, the momentum and weight decay
constants were kept at 0.9 and $5\times 10^{-4}$ respectively. Tables
\ref{table:hyp_cifar10} and \ref{table:hyp_cifar100} list the remaining
hyperparameters.  When a stopping accuracy of $100\%$ is listed, we mean that
the maximum number of epochs were used.

 \begin{table}[h] \centering \begin{tabular}{ccccc} \multicolumn{1}{c}{\bf
		CIFAR10}  &\multicolumn{1}{c}{\makecell{\bf SGD \\ \bf (small-batch)}} &
		\multicolumn{1}{c}{\bf \makecell{SGD \\ \bf(large-batch)}}  &
		\multicolumn{1}{c}{\bf \makecell{SWAP \\ (Phase 1)}} &
		\multicolumn{1}{c}{\bf \makecell{SWAP\\ (Phase 2)}} \\
 \hline  
 Batch-size & $512$ & $4096$ & $4096$ & $512$ \\ 
 Learning-rate Peak & $0.3$ & $1.2$ & $1.2$ & $0.12$ \\ 
 Maximum Epochs & $100$ & $150$ & $150$ & $30$ \\ 
 Warm-up Epochs & $30$ & $30$ & $30$ & $0$ \\ 
 GPUs used per model & $2$ & $8$ & $8$ & $1$  \\ 
 Stopping Accuracy (\%) & $100 $ & $100$ & $98$ & $100$  \\
 \hline \end{tabular} \caption{Hyperparameters obtained using tuning for CIFAR10} 
 \label{table:hyp_cifar10} \end{table}

 \begin{table}[h] \centering \begin{tabular}{ccccc} \multicolumn{1}{c}{\bf
		CIFAR100}  &\multicolumn{1}{c}{\makecell{\bf SGD \\ \bf (small-batch)}} &
		\multicolumn{1}{c}{\bf \makecell{SGD \\ \bf(large-batch)}}  &
		\multicolumn{1}{c}{\bf \makecell{SWAP \\ (Phase 1)}} &
		\multicolumn{1}{c}{\bf \makecell{SWAP\\ (Phase 2)}} \\
 \hline  
 Batch-size & $128$ & $2048$ & $2048$ & $128$ \\ 
 Learning-rate Peak & $0.2$ & $1.2$ & $1.2$ & $0.05$ \\
 Total Epochs & $150$ & $150$ & $150$ & $30$ \\  
 Warm-up Epochs & $60$ & $45$ & $45$ & $0$ \\ 
 GPUs used per model & $1$ & $8$ & $8$ & $1$  \\ 
 Stopping Accuracy (\%) & $100 $ & $100$ & $90$ & $100$  \\
 \hline \end{tabular} \caption{Hyperparameters obtained using tuning for CIFAR100} 
 \label{table:hyp_cifar100} \end{table}

\end{document}